\documentclass[10pt,twocolumn,letterpaper]{article}

\usepackage{wacv}
\usepackage{times}
\usepackage{epsfig}
\usepackage{subcaption}
\usepackage{graphicx}
\usepackage{caption}
\usepackage{amsmath}
\usepackage{amssymb}
\usepackage{adjustbox}
\usepackage{multirow}
\usepackage{multicol}
\usepackage{booktabs}
\usepackage{url}
\usepackage{array}
\usepackage[dvipsnames,table,xcdraw]{xcolor}
\usepackage[accsupp]{axessibility}  
\usepackage{pifont}

\definecolor{myorange}{rgb}{1, 0.647, 0}
\definecolor{myblue}{rgb}{.118, 0.565, 1}
\graphicspath{Figures/}

\def\wacvPaperID{1068} 

\wacvalgorithmstrack   

\wacvfinalcopy 

\ifwacvfinal
\usepackage[breaklinks=true,bookmarks=false]{hyperref}
\else
\usepackage[pagebackref=true,breaklinks=true,colorlinks,bookmarks=false]{hyperref}
\fi

\begin{document}

\title{BEFUnet: A Hybrid CNN-Transformer Architecture for Precise Medical Image Segmentation}

\renewcommand*{\thefootnote}{\fnsymbol{footnote}}

\author{Omid Nejati Manzari$\,$\footnotemark[1]$^{\;\,,1}$
\quad Javad Mirzapour Kaleybar$^{2}$
\quad Hooman Saadat$^{3}$
\quad Shahin Maleki$^{2}$
\\
${^1}$ School of Electrical Engineering, Iran University of Science and Technology, Tehran, Iran\\
${^2}$ Department of Computer Engineering, University College of Nabi Akram, Tabriz, Iran  \\
${^3}$ Department of Electrical Engineering, Qazvin Branch, Iran\\
}

\maketitle

\footnotetext[1]{Corresponding author \\ omid\_nejaty@alumni.iust.ac.ir}

\thispagestyle{empty}

\begin{abstract}
\vspace{-0.5em}

The accurate segmentation of medical images is critical for various healthcare applications. Convolutional neural networks (CNNs), especially Fully Convolutional Networks (FCNs) like U-Net, have shown remarkable success in medical image segmentation tasks. However, they have limitations in capturing global context and long-range relations, especially for objects with significant variations in shape, scale, and texture. While transformers have achieved state-of-the-art results in natural language processing and image recognition, they face challenges in medical image segmentation due to image locality and translational invariance issues.
To address these challenges, this paper proposes an innovative U-shaped network called BEFUnet, which enhances the fusion of body and edge information for precise medical image segmentation. The BEFUnet comprises three main modules, including a novel Local Cross-Attention Feature (LCAF) fusion module, a novel Double-Level Fusion (DLF) module, and dual-branch encoder.
The dual-branch encoder consists of an edge encoder and a body encoder. The edge encoder employs PDC blocks for effective edge information extraction, while the body encoder uses the Swin Transformer to capture semantic information with global attention.
The LCAF module efficiently fuses edge and body features by selectively performing local cross-attention on features that are spatially close between the two modalities. This local approach significantly reduces computational complexity compared to global cross-attention while ensuring accurate feature matching.
BEFUnet demonstrates superior performance over existing methods across various evaluation metrics on medical image segmentation datasets. The synergy of CNN and transformer architectures in BEFUnet excels in handling irregular and challenging boundaries, establishing it as a promising approach for advancing medical image segmentation tasks. Our code is publicly available at~\href{https://github.com/Omid-Nejati/BEFUnet}{\textcolor{magenta}{GitHub}}.
\end{abstract}

\vspace{-1.5em}
\section{Introduction}
Medical image segmentation is of utmost importance in computer vision, offering vital insights into anatomical regions for comprehensive analysis and aiding healthcare professionals in injury visualization, disease monitoring, and treatment planning. CNNs, particularly fully convolutional networks (FCNs) \cite{long2015fully} like U-Net \cite{ronneberger2015u}, have emerged as dominant methods, showing remarkable success in various medical applications such as cardiac, organ, and polyp segmentation. The field of automated medical image segmentation has gained significant attention for its potential to alleviate labor-intensive tasks for radiologists. FCNs and their variants, including SegNet \cite{badrinarayanan2017segnet}, U-Net \cite{ronneberger2015u}, DenseNet \cite{cai2020dense}, and DeepLab \cite{chen2017deeplab}, have played a crucial role in enhancing medical diagnostics and treatment by enabling accurate segmentation from MRI, CT scans, PET/CT scans, and other medical imaging modalities.

CNN-based approaches are widely favored for image segmentation due to their powerful local translation invariance, representation learning capabilities, and filter sharing properties. However, these methods have limitations in capturing long-range relations and explicit global context, especially for objects with large inter-patient variation in texture, scale, and shape. Various strategies like dilated convolution \cite{yu2015multi}, image pyramids \cite{zhao2017pyramid}, prior-guided methods \cite{cheng2020learning}, multi-scale fusion \cite{gu2022multi}, and self-attention mechanisms \cite{oktay2018attention} have been attempted to address these limitations in medical image segmentation but have shown weaknesses in extracting global context features. Among the successful architectures for medical image segmentation are the FCN and the U-Net, both utilizing convolutional layers on encoder and decoder modules. The U-Net \cite{ronneberger2015u}, with its skip connections, has proven to be versatile for segmentation tasks, but the convolution layer's locality restriction limits its representational power in capturing shape and structural information crucial for medical image segmentation. The restricted performance of CNN models in building long-range dependencies and global contexts further hinders their effectiveness in image segmentation. Addressing these challenges remains an ongoing pursuit in CNN architectures for medical image segmentation.

The recent success of transformers in Natural Language Processing (NLP) \cite{vaswani2017attention} has led to the development of vision transformers to overcome CNN limitations in image recognition tasks \cite{dosovitskiy2020image}. Vision transformers leverage multi-head self-attention (MSA) to establish long-range dependencies and capture global contexts. However, vanilla vision transformers demand large amounts of data to generalize and suffer from quadratic complexity. Various approaches like DeiT \cite{touvron2021training}, Swin Transformer, and pyramid vision transformer aim to address these limitations. While transformers have achieved state-of-the-art (SOTA) results in image classification and semantic segmentation, their high computation power requirements hinder real-time applications, such as radiotherapy. To address the limitations of CNN models, Vision Transformer (ViT) models utilize the MSA mechanism and achieve SOTA performance compared to convolution-based methods. Hybrid-transformer architectures integrate transformers with CNN-based networks, while pure transformers like Swin-Unet and others apply transformers to both the encoder and decoder for global feature representation at multiple levels. Despite transformers' ability to capture global dependencies, they struggle with image locality and translational invariance, affecting accurate segmentation of organ boundaries.

Recent advances in vision transformers emphasize the significance of multi-scale feature representations. DS-TransUNet \cite{lin2022ds}, CrossViT \cite{chen2021crossvit}, and HRViT \cite{gu2022multi} introduce dual-branch transformer architectures to extract contextual information, enhance fine-grained features, and improve semantic segmentation. In medical image processing, researchers explore multi-branch architectures to address challenges, such as Valanarasu et al. \cite{valanarasu2020kiu} using undercomplete and overcomplete encoders, Lin et al. \cite{lin2022ds} employing a dual-branch Swin Transformer, and Zhu et al. \cite{zhu2023brain} leveraging different branches for processing various data modalities and achieving superior performance. These multi-branch models show promise in tackling complex features in medical image processing.

Despite the ability of vision transformers to capture global contextual representations, their self-attention mechanism can overlook low-level features. Hybrid CNN-transformer approaches, such as TransUnet \cite{chen2021transunet} and LeVit-Unet \cite{xu2021levit}, have been proposed to address this issue by combining the locality of CNNs with the long-range dependency of transformers to encode both global and local features in medical image segmentation. However, these approaches face challenges in effectively combining high-level and low-level features while maintaining feature consistency and utilizing multi-scale information produced by the hierarchical encoder properly. Furthermore, many existing deep learning-based methods for medical image segmentation tend to focus solely on body features, neglecting the significance of edge information. Some studies, like Kuang et al. \cite{kuang2021bea} and Yang \cite{yang2023cswin}, attempted to incorporate edge features by separating them from body features or using them as additional constraints. However, these approaches did not fully exploit the potential of edge features in medical image segmentation.

In this paper, we introduce an innovative network called Body and Edge Fusion Unet (BEFUnet), which aims to achieve precise medical image segmentation by enhancing the fusion of edge and body information. The proposed BEFUnet consists of three key modules: a dual-branch encoder, a Double-Level Fusion (DLF) module, and a Local Cross-Attention Feature fusion module (LCAF). The dual-branch encoder is designed to simultaneously extract edge and body information using a lightweight CNNs branch with pixel-wise convolution and a hierarchical Transformer branch based on SwinTransformer. The LCAF effectively fuses cross-modal features by considering features that are closely located in position between the two branches, leading to improved accuracy while reducing computational complexity. Moreover, the novel DLF module is a multi-scale vision transformer that employs a cross-attention mechanism to fuse two obtained feature maps. The proposed BEFUnet not only addresses the aforementioned issues but also outperforms other methods in terms of various evaluation metrics. Our main contributions:

\noindent $\bullet$ A novel hybrid method that combines the edge local semantic information of CNN with the body contextual interactions of the transformer, resulting in enhanced integration of complementary features. This approach proves to be particularly advantageous in handling irregular and challenging boundaries in medical image segmentation.

\noindent $\bullet$ We introduce a Double-Level Fusion module, which effectively fuses coarse and fine-grained feature representations.

\noindent $\bullet$ Our experimental results underscore the remarkable effectiveness of BEFUnet, as we meticulously trained and evaluated our model on three distinct medical image segmentation datasets. In our comprehensive comparison, BEFUnet consistently outperforms a diverse set of state-of-the-art models, affirming its robustness and superiority in handling various segmentation challenges across different datasets.




\section{Related Works}
\vspace{-0.5em}
\subsection{CNN-based Segmentation Networks}
\vspace{-0.5em}
In early medical image segmentation, traditional machine learning algorithms \cite{tsai2003shape} were widely used. However, the landscape changed with the emergence of deep CNNs, U-Net \cite{ronneberger2015u}, and its variants became powerful alternatives. The simplicity and superior performance of U-shaped structures have made them popular choices for both 2D and 3D medical image segmentation scenarios.
In the realm of medical image segmentation, the focus primarily shifted toward the application of CNNs and their variations to achieve accurate and efficient segmentation results. Researchers introduced various methods to address the limitations of vanilla FCNs \cite{long2015fully}. These methods include fusing the output of different layers, using dilated convolutions \cite{yu2015multi}, and employing context modeling \cite{zhao2017pyramid}. Furthermore, U-shaped encoder-decoder structures like U-Net and its variants garnered significant attention as they offered further enhancements in performance for medical imaging tasks.
Among these architectures, U-Net \cite{ronneberger2015u} has become the standard choice for medical image analysis. It consists of a symmetric encoder and decoder network with skip connections, which prove highly effective. Building upon this success, several U-Net-like architectures have been proposed, such as Res-UNet \cite{xiao2018weighted}, Dense-UNet \cite{cai2020dense}, Kiu-Net \cite{valanarasu2020kiu}, and U-net++ \cite{zhou2018unet++}. Each of these variants offers improvements and adaptations to cater to different medical imaging tasks. For example, U-net++ introduces dense skip connections between modules, leading to improved results. As a result, these structures have demonstrated outstanding performance across various medical domains.
Despite the remarkable progress made with CNN-based methods in medical image segmentation, there are still limitations in terms of segmentation accuracy and efficiency due to the inherent locality of convolution operations and complex data access patterns. Efforts continue to address these challenges and further improve the capabilities of medical image segmentation models.


\vspace{-0.25em}
\subsection{Vision Transformers}
\vspace{-0.5em}

In recent years, Transformer-based models have demonstrated remarkable success across various domains, particularly in natural language processing. These models have achieved SOTA performance in tasks such as machine translation \cite{vaswani2017attention}. Moreover, their application has extended to vision tasks with the introduction of the Vision Transformer (ViT) \cite{dosovitskiy2020image}. The ViT has showcased an impressive speed-accuracy trade-off for image recognition; however, it requires pre-training on large datasets. To address this limitation, efforts have been made to enhance ViT's performance on ImageNet, resulting in Deit \cite{touvron2021training}, which incorporates training strategies for improved outcomes.
Additionally, another noteworthy vision Transformer is the Swin Transformer \cite{liu2021swin}, a hierarchical model that serves as an efficient and effective vision backbone. The Swin Transformer has attained SOTA performance in image classification, object detection, and semantic segmentation tasks.

\vspace{-0.25em}
\subsection{Transformers for Medical Image Segmentation}
\vspace{-0.5em}

In the context of medical segmentation, researchers have explored the potential of Transformer-based models \cite{xiao2023transformers, zhao2023m, gao2022data, ruan2023ege}. Several variants have been introduced, such as MedT \cite{valanarasu2021medical}, which integrates gated axial transformer layers into existing architectures, and TransUNet \cite{chen2021transunet}, which combines Transformers and CNNs to leverage their respective advantages. Moreover, Swin-UNet \cite{cao2021swin} has adopted pure transformers within a U-shaped encoder-decoder architecture for global semantic feature learning. These Transformer-based approaches show promise in overcoming the limitations of CNN models, which struggle with long-range dependency modeling due to their restricted receptive fields. By incorporating both global and local information, these hybrid models aim to improve feature extraction for 3D and 2D segmentation of medical images \cite{zhu2023brain, bai2022transfusion, li2022deepfusion}.

To further enhance the fusion of edge and body information, the proposed BEFUnet architecture stands out. BEFUnet employs a novel transformer-based fusing scheme, ensuring feature consistency and richness in 2D medical image segmentation. This approach addresses the limitations of simple feature-fusing mechanisms found in other hybrid models and guarantees more effective feature maps that contain rich information, leading to improved segmentation results.

\begin{figure*}[ht]
	\centering
	\includegraphics[width=0.75\textwidth]{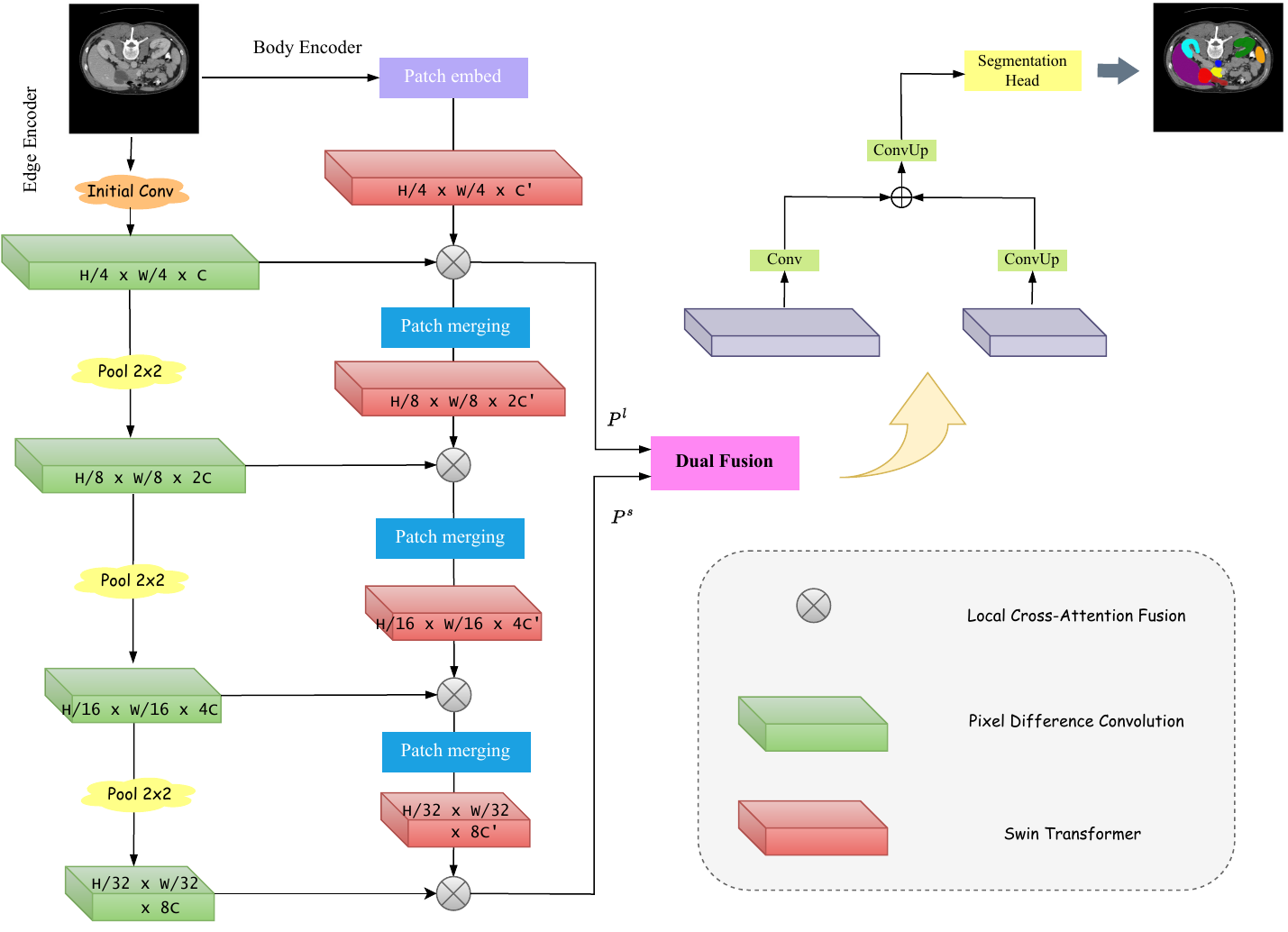}
 \vspace{-1em}
	\caption{\textbf{The structure of our proposed BEFUnet.} BEFUnet consists of a dual-encoder that simultaneously extracts edge and body features. Moreover, it incorporates an efficient fusion module called LCAF, which facilitates the merging of edge and body features. Additionally, there is a DLF module integrated into the skip connection of the encoder-decoder structure to fully integrate features from adjacent scales.}
	 \label{fig:mainfig}
\end{figure*}


\section{Methodology}

The framework of BEFUnet is depicted in Figure \ref{fig:mainfig}. It consists of a dual-branch encoder, a cross-attention feature fusion module called LCAF, and a DLF module, which effectively fuses coarse and fine-grained feature representations. In the upcoming sections, we will provide a detailed description of these components.

\subsection{Dual-branch encoder}

The dual-branch encoder comprises a body encoder and one edge encoder. The edge encoder utilizes CNNs architecture with Pixel Different Convolution (PDC) \cite{su2021pixel} blocks for extracting edge information effectively. On the other hand, the body encoder employs Transformer architecture for capturing semantic information using global attention.

\subsubsection{Edge encoder}

The extraction of edge features is crucial for image segmentation, and to address the issue of inadequate edge information extraction by conventional medical segmentation networks, a dedicated edge detection branch is incorporated. As illustrated in Figure \ref{fig:mainfig}, the edge encoder consists of four stages, each containing four PDC blocks for feature detection. Max-pooling layers are utilized for downsampling the feature maps between stages to obtain hierarchical features. The first stage expands the original 3-channel image to C channels and reduces the feature maps' size to 1/4, ensuring compatibility with the body encoder's output size.

The PDC block consists of a depth-wise convolution layer, a ReLU layer, and a convolution layer with a kernel size of 1. A residual connection is also included to aid in model training. Traditional convolutional networks face difficulties in extracting edge-related information due to the lack of explicitly encoded gradient information. To overcome this, PDC is introduced, which integrates gradient information into convolutional operations and enhances edge features.

Vanilla convolution calculates the weighted sum of pixel values within the convolution kernel, while PDC calculates the weighted sum of differences between pixel values within the kernel. The expressions for vanilla convolution and PDC are given by Equations (1) and (2), respectively:

\begin{equation}
y=f(x,\theta)=\sum_{i=1}^{k\times k}w_i\cdot x_i \quad (\text{vanilla convolution})
\end{equation}

\begin{equation}
y=f(\nabla x,\theta)=\sum_{(x_i,{x}'_i)\in P}w_i\cdot (x_i-x_i') \quad (\text{PDC})
\end{equation}

Here, $w_i$ represents the weights in the k × k convolution kernel, $x_i$ and ${x}'_i$ denote the pixels covered by the kernel, and $P$ represents the collection of selected pixel pairs in the local area covered by the kernel.

To further enhance edge extraction, a supervision strategy \cite{xie2015holistically} is employed, where an edge map is generated for the output feature of each stage, and the loss is computed between the generated edge maps and the ground truth.

\subsubsection{Body encoder}

The body encoder utilizes the Transformer architecture, specifically the Swin Transformer \cite{liu2021swin}, for encoding high-level feature representations with global long-range modeling. The Swin Transformer employs a sliding window mechanism to construct hierarchical features and is well-suited for segmenting organs with irregular shapes.

Figure \ref{fig:mainfig} illustrates the overall structure of the Swin Transformer in red blocks, consisting of four stages. The first stage performs feature encoding on the original image using a patch embedding layer and two Swin Transformer blocks. The image is divided into patches of size $P\times P$, reshaped into 1D vectors, and projected to a C-dimensional space. Positional parameters are added to encode positional information, followed by passing the sequence through the Swin Transformer blocks. The subsequent stages downsample the feature map and extract higher-level features using patch merging and Swin Transformer blocks.

Each stage of the Swin Transformer comprises alternating arrangements of two different types of blocks. The first block consists of Layer Normalization (LN), Window-based Multi-head Self-Attention (W-MSA), Multi-Layer Perceptron (MLP), and residual connections. The second block is similar, but it replaces W-MSA with Shifted Window-based Multi-head Self-Attention (SW-MSA). The operations within these blocks can be expressed using Equations (3)-(6).

\begin{flalign}
&\  \hat{z}^l=W\mbox{-}MSA(LN(z^{l-1}))+z^{l-1}  &\\
&\  z^{l}=MLP(LN(\hat{z}^{l}))+\hat{z}^{l} &\\
&\  \hat{z}^{l+1}=SW\mbox{-}MSA(LN(z^{l}))+z^{l}  &\\
&\   z^{l+1}=MLP(LN(\hat{z}^{l+1}))+\hat{z}^{l+1}  &
\end{flalign}

Between each stage, patch merging is performed to downsample the feature map and gather contextual features. It merges adjacent $2\times 2$ patches into a larger patch, reducing the number of patches and concatenating their dimensions to minimize information loss. Patch merging downsamples the features by a factor of 2. The output feature size of each stage is $\frac{H}{4}\times \frac{W}{4}\times C$, $\frac{H}{8}\times \frac{W}{8}\times 2C$, $\frac{H}{16}\times \frac{W}{16}\times 4C$, and $\frac{H}{32}\times \frac{W}{32}\times 8C$, respectively, assuming the input image size is $H\times W\times 3$.

The extracted edge and body features are then fed into the LCAF module for further fusion.

\begin{figure}[htbp]
		\centering
            {\includegraphics[width=0.8\columnwidth]{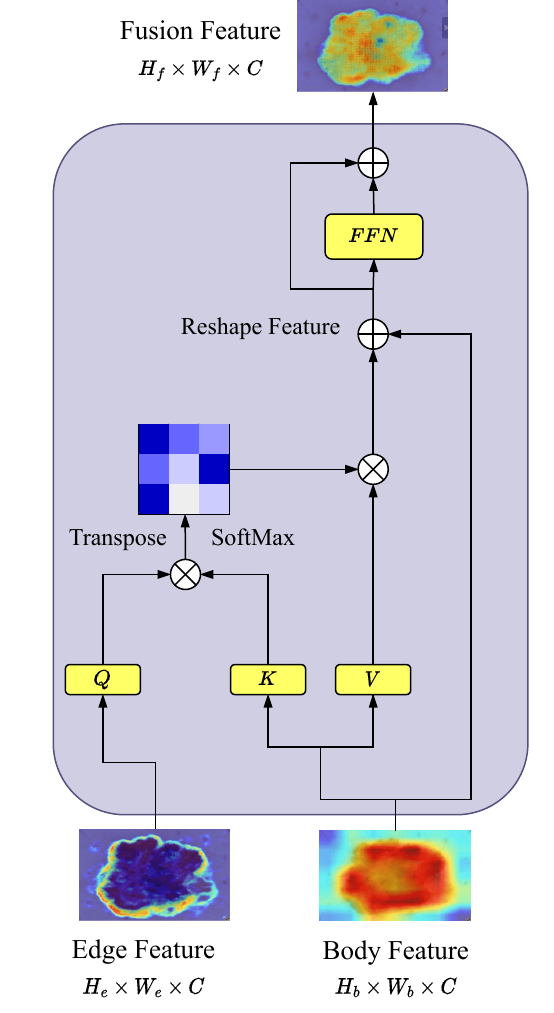}}	
		\caption{A block diagram of the LCAF.}
        \label{Fig:LCAF}
	\end{figure}

\subsection{LCAF module}

The Local Cross-Attention Fusion (LCAF) module is designed to fuse edge and body features accurately and efficiently by selectively performing cross-attention on features that are spatially close between the two image modalities. This approach overcomes the limitations of conventional cross-attention methods, which require storing global information for each patch and cannot leverage the one-to-one correspondence of pixels between modalities for accurate matching and comprehensive fusion.

Figure \ref{Fig:LCAF} illustrates the structure of the LCAF module. It utilizes a transformer block with local cross-attention to modulate features in close proximity. The input edge and body features are divided into local regions, projected into query, key, and value vectors, and attention scores are calculated through dot product attention. The resulting vectors are weighted and summed to obtain the fused feature representation. The LCAF module also incorporates a multi-head operation to project vectors onto distinct subspaces and compute cross-attention scores in diverse subspaces. The output of the LCAF module is obtained by applying feed-forward networks (FFN) to the modulated features, as described in Equation (9). Residual connections are used to maintain feature integrity.

\begin{equation}
LCA(X_f) = softmax(\frac{X_{edge}W_Q(X_{body}W_K)^T}{\sqrt{d_k} })X_{body}W_v
\end{equation}

\begin{equation}
M\mbox{-}LCA(X_f) = X_f + Concat[LCA(X_f)_1, ... , LCA(X_f)_h]W_o
\end{equation}

\begin{equation}
X_f=FFN(M\mbox{-}LCA(X_f)) + M\mbox{-}LCA(X_f)
\end{equation}

\noindent In this context, $X_{body}$, $X_{edge}$, and $X_f$ refer to features corresponding to the edge modality, body modality, and fused modality, respectively. The learnable matrices $W_k$, $W_Q$, $W_V$, and $W_o$ represent the key matrix, query matrix, value matrix, and a matrix for output, respectively.
Compared to conventional global cross-attention, LCAF significantly reduces computational complexity by performing selective local cross-attention. The computational complexities of global cross-attention (GCA) and Local Cross-Attention (LCA) are given by Equations (10) and (11), respectively, assuming $h$ and $w$ patches in the features and $h_l$ and $w_l$ patches in the LCA window.

\begin{align}
&\  \Omega (M\mbox{-}GCA) = 4hwC^2 + 2(hw)^2C  &\\
&\  \Omega (M\mbox{-}LCA) = 4hwC^2 + 2h_lh_w\cdot hwC &
\end{align}

The LCAF module outputs the fused features, which are then fed into the next stage.

\begin{figure*}[ht]
	\centering
	\includegraphics[width=0.95\textwidth]{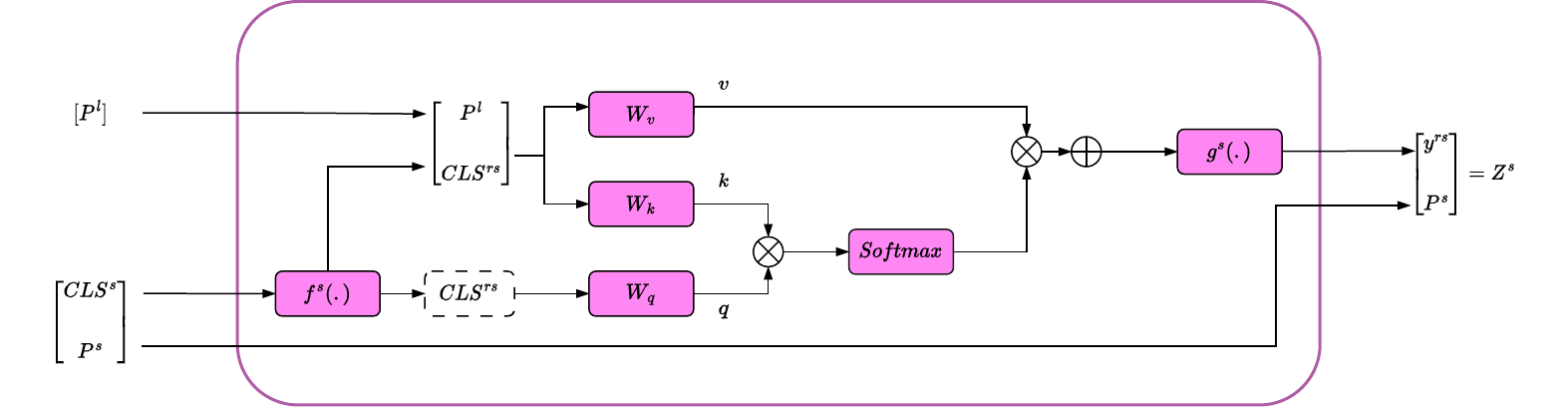}
 \vspace{-1em}
	\caption{The Cross Attention process entails several steps. Initially, the class token of the small level, denoted as $CLS^s$, is projected for dimension alignment and then appended to $P^l$. The resulting embedding operates as both the key and value. Moreover, the query is made using $CLS'^s$. Subsequently, attention computation and back projection are performed to obtain $Z^s$. Noteworthily, this process can also be extended to the large level.}
	 \label{fig:DLF}
\end{figure*}

\subsubsection{Double-Level Fusion Module (DLF)}

The primary obstacle lies in effectively merging CNN and Swin Transformer level features while maintaining feature consistency. A direct approach would be to input the sum of CNN levels and their corresponding Swin Transformer levels into a decoder to obtain the segmentation map. However, this approach fails to ensure feature consistency between the levels, resulting in subpar performance. Therefore, we propose a novel module called DLF that addresses this issue by taking the smallest ($P^{s}$) and largest ($P^{l}$) levels as inputs and employing a cross-attention mechanism to fuse information across scales.

Typically, shallow levels contain more precise localization information, while deeper levels carry more semantic information that is better suited for the decoder. To balance the trade-off between computational cost and the marginal effect of the middle-level feature map on model accuracy, we opted not to include the middle level in the feature fusion process, thus saving computational resources. Consequently, we encourage representation by multi-scaling the shallowest ($P^s$) and the last ($P^l$) levels while preserving localization information.

In the proposed DLF module, the class token plays a significant role as it summarizes all the information from the input features. Each level is assigned a class token derived from global average pooling (GAP) over the level's norm. The class tokens are obtained as follows:

\begin{equation}
\vspace{-0.25em}
\begin{aligned}
CLS^{s} &= GAP(Norm (P^{s}))
\\
CLS^{l} &= GAP(Norm (P^{l}))
\end{aligned}
\end{equation}
where $CLS^{s}$ $\in R^{4D' \times 1}$ and $CLS^{l}$ $\in R^{D'\times 1}$. The class tokens are then concatenated with the respective level embeddings before being passed into the transformer encoders. The small level is followed by $S$ transformer encoders, and the large level is followed by $L$ transformer encoders to compute global self-attention. Notably, a learnable position embedding is also added to each token of both levels to incorporate position information into the transformer encoders' learning process.

After passing the embeddings through the transformer encoders, the features of each level are fused using the cross-attention module. Prior to fusion, the class tokens of the two levels are swapped, meaning the class token of one level is concatenated with the tokens of the other level. Each new embedding is then individually fed through the module for fusion and finally back-projected to its original level. This interaction with tokens from the other level allows the class tokens to share rich information across levels.

In particular, for the small level, this interaction is illustrated in Fig. \ref{fig:DLF}. $f^s(.)$ first projects $CLS^s$ to the dimensionality of $P^l$, and the resulting output is denoted as $CLS'^s$. $CLS'^s$ is concatenated with $P^l$ to serve as the key and value for the cross-attention computation, while also independently acting as the query. Since only the class token is queried, the cross-attention mechanism operates in linear time. The final output $Z^s$ can be mathematically represented as follows:
\begin{align}
&y^{s} = f^{s}(CLS^s)+MCA(LN([f^s(CLS^s)\parallel P^l]))
\nonumber\\
&Z^s=[P^s \parallel g^s(y^s)]
\label{eq.MCA}
\end{align}

\subsection{Loss function}

Since the encoder is designed to detect both edge and body features simultaneously, we employ two loss components: $L_{edge}$ and $L_{body}$.

\subsubsection{Edge supervision loss}

For edge detection, a significant majority of samples are negative, requiring the use of an annotator-robust loss \cite{liu2017richer}.

We apply this loss function to the edge maps generated by each stage of the encoder. For the $i$-th pixel in the $j$-th edge map, denoted as $y^j_i$, the loss is computed according to the following conditions:

\begin{equation}
l^j_i =
\begin{cases}
\alpha \cdot \log(1-y^j_i) & \text{if } y_i = 0 \\
0 & \text{if } 0 < y_i < \eta \\
\beta \cdot \log y^j_i & \text{otherwise}
\end{cases}
\end{equation}

Here, $y^j_i$ represents the predicted value of the $i$-th pixel in the $j$-th edge map, and $\eta$ is a predefined threshold. Pixels that are annotated as positive by annotators with a proportion smaller than $\eta$ are considered negative samples. The proportion of negative samples in the dataset is denoted as $\beta$. Additionally, we define $\alpha = \lambda \cdot (1-\beta)$, where $\lambda$ is a hyperparameter used to balance positive and negative samples. The overall $L_{edge}$ loss is obtained by summing the losses of each pixel:

\begin{equation}
L_{edge} = \sum_{i,j} l^j_i
\end{equation}

\subsubsection{Body supervision loss}

Medical image segmentation faces a severe class imbalance problem, which we address by employing a combination of Binary Cross-Entropy Loss and Dice Loss \cite{milletari2016v} as the body loss.

Binary Cross-Entropy Loss measures pixel-level prediction errors and is suitable for most semantic segmentation scenarios. It can be expressed as follows:

\begin{equation}
L_{Bce} = - \sum_{i=0}^N [(1-\hat{y}_i) \log(1-y_i) + \hat{y}_i \log(y_i)]
\end{equation}

Dice Loss is a widely used loss function in medical image segmentation, effectively handling imbalanced samples. It is given by the following formula:

\begin{equation}
L_{Dice} = 1 - 2 \times \frac{2 \sum_{i=0}^N y_i \hat{y}i}{\sum{i=0}^N (y_i + \hat{y}_i)}
\end{equation}

The $L_{body}$ loss can be expressed as a combination of Binary Cross-Entropy Loss and Dice Loss:

\begin{equation}
L_{body} = \lambda_1 L_{Bce} + \lambda_2 L_{Dice}
\end{equation}

Finally, combining $L_{edge}$ and $L_{body}$ with a weighted hyperparameter $\gamma$, we obtain the final loss $L$:

\begin{equation}
L = L_{body} + \gamma \cdot L_{edge}
\end{equation}

\begin{table*}[!ht]
    \centering
    \caption{Comparison results of the proposed method on the \textit{Synapse} dataset. \textcolor{blue}{Blue} indicates the best result, and \textcolor{red}{red} displays the~second-best.}
    \label{Tab:Synapse}
    \resizebox{\textwidth}{!}{
    \begin{tabular}{l|cc|cccccccc}
    \toprule
    \textbf{Methods} & \textbf{DSC~$\uparrow$} & \textbf{HD~$\downarrow$} & \textbf{Aorta} & \textbf{Gallbladder} & \textbf{Kidney(L)} & \textbf{Kidney(R)} & \textbf{Liver} & \textbf{Pancreas} & \textbf{Spleen} & \textbf{Stomach} \\
    \midrule
    DARR \cite{fu2020domain} & 69.77 & - & 74.74 & 53.77 & 72.31 & 73.24 & 94.08 & 54.18 & 89.90 & 45.96
    \\
    V-Net \cite{milletari2016v}                   & 68.81 & - & 75.34 & 51.87 & 77.10 & \textcolor{red}{80.75} & 87.84 & 40.05 & 80.56 & 56.98
    \\
    U-Net \cite{ronneberger2015u} & 76.85 & 39.70 & \textcolor{red}{89.07} & \textcolor{red}{69.72} & 77.77 & 68.60 & 93.43 & 53.98 & 86.67 & 75.58
    \\
    R50 U-Net \cite{chen2021transunet} & 74.68 & 36.87 & 87.74 & 63.66 & 80.60 & 78.19 & 93.74 & 56.90 & 85.87 & 74.16
    \\
    Att-UNet \cite{schlemper2019attention}  & 77.77 & 36.02 & \textcolor{blue}{89.55} & 68.88 & 77.98 & 71.11 & 93.57 & 58.04 & 87.30 & 75.75
    \\
    R50 Att-UNet \cite{chen2021transunet}  & 75.57 & 36.97 & 55.92 & 63.91 & 79.20 & 72.71 & 93.56 & 49.37 & 87.19 & 74.95
    \\
    R50 ViT \cite{chen2021transunet}  & 71.29 & 32.87 & 73.73 & 55.13 & 75.80 & 72.20 & 91.51 & 45.99 & 81.99 & 73.95
    \\
    Swin-Unet \cite{cao2021swin} & 79.13 & 21.55 & 85.47 & 66.53 & 83.28 & 79.61 & 94.29 & 56.58 & 90.66 & 76.60
    \\
    TransUnet \cite{chen2021transunet} & 77.48 & 31.69 & 87.23 & 63.13 & 81.87 & 77.02 & 94.08 & 55.86 & 85.08 & 75.62
    \\
    LeVit-Unet \cite{xu2021levit} & 78.53 & \textcolor{red}{16.84} & 78.53 & 62.23 & \textcolor{red}{84.61} & 80.25 & 93.11 & 59.07 & 88.86 & 72.76
    \\
    DeepLabv3+ (CNN) \cite{chen2018encoder} & 77.63 & 39.95 & 88.04 & 66.51 & 82.76 & 74.21 & 91.23 & 58.32 & 87.43 & 73.53
    \\
    M2SNet \cite{zhao2023m} & 79.70 & 21.39 & 85.48 & 66.89 & 83.03 & 79.12 & 94.40 & 56.26 & 90.11 & 76.97
    \\
    UTNetV2 \cite{gao2022multi} & 78.35 & 20.88 & 87.57 & 64.75 & 84.21 & 80.20 & \textcolor{red}{94.77} & 58.93 & 89.23 & 75.46
    \\
    MISSFormer \cite{huang2021missformer} & \textcolor{red}{79.74} & 19.65 & 85.31 & 66.47 & 83.37 & \textcolor{blue}{81.65} & 94.52 & \textcolor{blue}{63.49} & \textcolor{red}{91.51} & \textcolor{red}{79.63}
    \\
    \midrule
    \textbf{BEFUnet} & \textcolor{blue}{80.47} &  \textcolor{blue}{16.26} & 87.03 & \textcolor{blue}{73.89} & \textcolor{blue}{85.23} & {80.47} & \textcolor{blue}{95.49} & \textcolor{red}{60.31} & \textcolor{blue}{91.56} & \textcolor{blue}{81.54}\\
    \bottomrule
    \end{tabular}
    }
\end{table*}

\begin{figure*}[ht]
	\centering
	\includegraphics[width=0.7\textwidth]{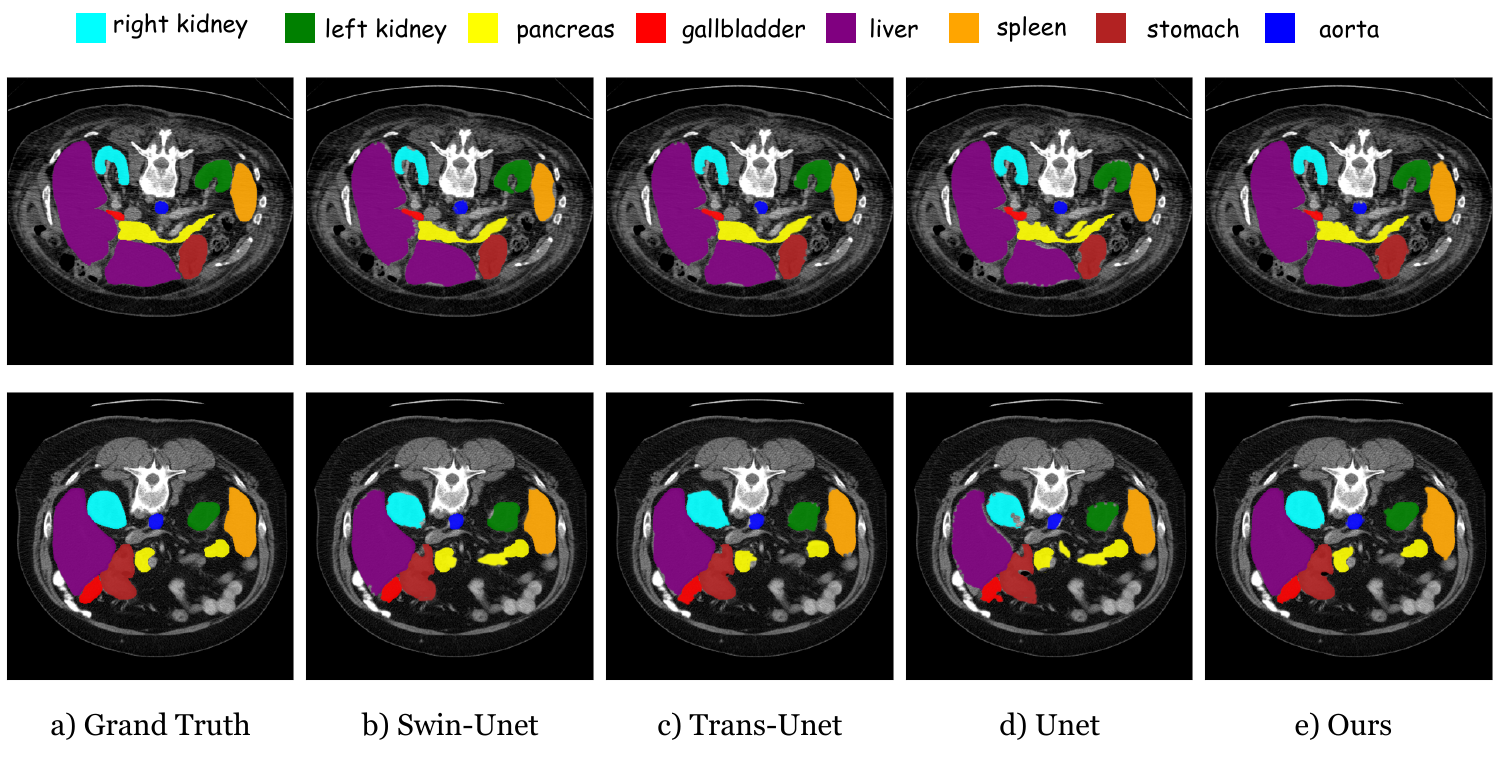}
 \vspace{-1em}
	\caption{Segmentation results of the proposed method on the \textit{Synapse} dataset. The red rectangles identify organ regions where the superiority of our proposed method can be clearly seen.}
	\label{fig:synapseviz}
\end{figure*}

\vspace{-0.5em}
\section{Experiments}

\vspace{-0.5em}
\subsection{Dataset}
\vspace{-0.5em}
\textbf{Synapse Multi-Organ Segmentation:}
To assess the performance of HiFormer, we conducted evaluations using the synapse multi-organ segmentation dataset \cite{synapse2015ct}. This dataset consists of 30 cases containing 3779 axial abdominal clinical CT images. Each CT volume comprises 85 to 198 slices of 512x512 pixels, with a voxel spatial resolution ranging from $([0.54 \sim 0.54] \times [0.98 \sim 0.98] \times [2.5 \sim 5.0]) mm^3$.

\textbf{Multiple Mylomia Segmentation:}
We evaluated our methodology on the multiple myeloma cell segmentation grand challenges provided by SegPC 2021 \cite{segpc2021,gupta2018pcseg}. The challenge dataset includes a training set with 290 samples, as well as separate validation and test sets with 200 and 277 samples, respectively.

\textbf{Skin Lesion Segmentation:}
Extensive experiments were conducted on skin lesion segmentation datasets, including the ISIC 2017 dataset \cite{codella2018skin}. This dataset consists of 2000 dermoscopic images for training, 150 for validation, and 600 for testing. Additionally, we used the ISIC 2018 dataset \cite{codella2019skin} and followed the methodology described in previous literature \cite{wu2022fat} to split the dataset into train, validation, and test sets. We also incorporated the $\mathrm{PH}^{2}$ dataset \cite{mendoncca2013ph}, which is a dermoscopic image database designed for segmentation and classification tasks. The $\mathrm{PH}^{2}$ dataset contains 200 dermoscopic images, including 160 nevi and 40 melanomas.

\subsection{Implementation Details}
\vspace{-0.5em}
The proposed model was implemented using the PyTorch framework and all experiments were conducted on an Nvidia Titan RTX 24G GPU. For improved computational efficiency, the resolution of all training, validation, and test images was set to $224\times 224$. To achieve better model initialization, we employed PiDiNet \cite{su2021pixel} as the pre-trained model for the edge encoder and swin-tiny \cite{liu2021swin} for the body encoder.

During training, the proposed model utilized the AdamW optimizer with a weight decay of 0.01. The initial learning rate was set to 0.01, and the ReduceLROnPlateau algorithm was employed for learning rate scheduling. A batch size of 24 and 80 training epochs were chosen. The weight parameters $\lambda _1$, $\lambda _2$, and $\gamma$ were set to 0.6, 0.4, and 0.2, respectively.

Furthermore, we utilize ReduceLROnPlateau for learning rate optimization. All experiments are performed on four datasets using identical train, validation, and test data. Additionally, we train other SOTA models with default parameters, while simultaneously employing a pretrained ViT model during the training of TransUNet and LeViT-UNet. On the other hand, the remaining models are trained from scratch.

%

\section{Evaluation Results}
We assessed the accuracy performance of our proposed method using commonly employed metrics, such as Sensitivity (SE), Specificity (SP), Accuracy (ACC), Intersection over Union (IoU), Dice coefficient (Dice), and Hausdorff Distance (HD). To ensure a fair and unbiased comparison, we evaluated BEFUnet against CNN and transformer-based methods, as well as models combining elements of both approaches.
\subsection{Results of Synapse Multi-Organ Segmentation}
Table \ref{Tab:Synapse} presents a comparison of our proposed BEFUnet with previous SOTA methods on the Synapse multi-organ CT dataset. The experimental results demonstrate that our Unet-like transformer method achieves outstanding performance, with a segmentation accuracy of 80.47\% (DSC) and 16.26\% (HD). Our algorithm show significant improvement in the DSC evaluation metric compared to Swin-Unet \cite{cao2021swin} and the recently introduced LeVit-Unet \cite{xu2021levit}, we achieved noteworthy accuracy enhancements of approximately 1\% and 5\% on the HD evaluation metric, indicating superior edge prediction capabilities. Notably, BEFUnet consistently outperforms existing literature in the segmentation of most organs, especially for the kidney, stomach, and liver.

Figure \ref{fig:synapseviz} displays the segmentation results of various methods on the Synapse dataset, revealing that these approaches generally struggle with edge region segmentation of organs. Our proposed BEFUnet, on the other hand, effectively addresses this challenge by leveraging a powerful edge encoder, resulting in enhanced segmentation, particularly in complex background scenarios.
\begin{table}[t]\footnotesize
\renewcommand{\arraystretch}{1.5}
    \centering
    \caption{Performance evaluation on the SegPC challenge 2021}
    \label{Tab:SegPC}
    \resizebox{0.95\columnwidth}{!}{%
    \begin{tabular}{llllll}
      \toprule
      Method & Accuracy & Precision & Recall & F1-score & mIoU\\
      \toprule
      R2U-Net \cite{alom2018recurrent}  & 0.933 & 0.852 & 0.831 & 0.834 & 0.744\\
      U-Net \cite{ronneberger2015u} & 0.939 & 0.842 & 0.879 & 0.855 & 0.766\\
      Unet++ \cite{zhou2018unet++} & 0.942& 0.855 & 0.876 & 0.857 & 0.770\\
      DoubleU-Net \cite{jha2020doubleu} & 0.937 & 0.833 & 0.896 & 0.858 & 0.763\\
      Att-UNet \cite{oktay2018attention} & 0.940 & 0.845 & 0.866 & 0.849 & 0.757\\
      ResUNet++ \cite{jha2019resunet++}  & 0.934 & 0.838 & 0.858 & 0.840 & 0.736\\
      UNet3+ \cite{huang2020unet} & 0.939 & 0.848 & 0.866 & 0.852 & 0.766\\
      LeViT-UNet \cite{xu2021levit} & 0.939 & 0.850 & 0.837 & 0.837 & 0.738\\
      TransUNet \cite{chen2021transunet} & 0.939 & 0.822 & 0.869 & 0.838 & 0.741\\
      M2SNet \cite{zhao2023m} & 0.945 & 0.839 & 0.866 & 0.848 & 0.751\\
      UTNetV2 \cite{gao2022multi} & 0.926 & 0.848 & 0.845 & 0.842 & 0.743\\
      MISSFormer \cite{huang2021missformer} & 0.947 & 0.865 & 0.871 & 0.851 & 0.752\\
      \midrule
      BEFUnet & \textbf{0.951} & \textbf{0.873} & \textbf{0.908} & \textbf{0.871} & \textbf{0.776}\\
      \bottomrule
    \end{tabular} %
    }
    \vspace{-1em}
\end{table}

\begin{figure}[htb]
\centering
    \includegraphics[width=0.95\columnwidth]{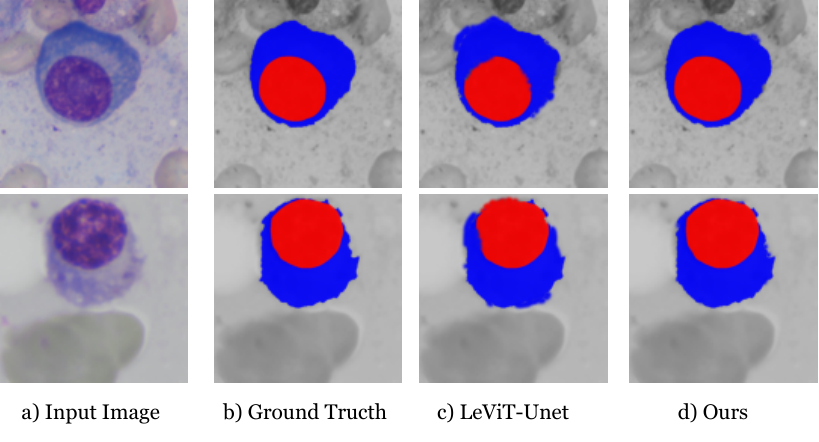}
    \caption{Visual representation of the proposed method on the \textit{SegPC} cell segmentation dataset.}
    \label{fig:segpc}
\end{figure}
\subsection{Results of Multiple Mylomia Segmentation}
In the domain of medical image analysis, it is common to encounter medical images that contain multiple classes of objects requiring segmentation. To address this requirement, we evaluated all models using the SegPC-2021 dataset, which comprises two distinct types of cells. The quantitative results can be found in Table \ref{Tab:SegPC}. Our proposed model outperforms other SOTA approaches, achieving a higher Accuracy score of 95.1.6\%, which represents a significant improvement of 1.2\% over Swin-Unet and a 1.3\% increase in F1-score compared to the DoubleU-Net architecture.

Furthermore, in Figure \ref{fig:segpc}, we provide visual illustrations of the segmentation outputs produced by our proposed model and LeVit-Unet. As demonstrated, our predictions align closely with the provided ground truth masks, indicating the effectiveness of our approach. One of the notable advantages of BEFUnet is its capability to model multi-scale representations. This property enables it to effectively suppress background noise, which is particularly valuable when dealing with datasets featuring highly overlapped backgrounds, such as SegPC. In summary, BEFUnet surpasses CNN-based methods that rely solely on local information modeling, as well as transformer-based counterparts, which exhibit poorer performance in this context.


%

\begin{table}[htbp]\footnotesize
\renewcommand{\arraystretch}{1.5}
  \begin{center}
    \caption{Results of BEFUnet on ISIC2017 dataset compared with the latest methods.}
    \label{Tab:ISIC}
    \begin{tabular*}{\hsize}{@{}@{\extracolsep{\fill}}l l l l l l@{}}
    \hline
      Method & Dice & SE &SP &ACC&IoU\\
      \hline
      U-Net \cite{ronneberger2015u} & 0.783 & 0.806 & 0.954 & 0.933 & 0.696\\
      Att U-Net \cite{oktay2018attention} & 0.808 & 0.800&0.978&0.915&0.717\\
      UNet++ \cite{zhou2018unet++}& 0.832 & 0.830 & 0.965 & 0.925 &0.743\\
      FocusNet \cite{kaul2019focusnet}& 0.832 & 0.767&0.980&0.921&0.756\\
      DAGAN \cite{lei2020skin}& 0.859 & 0.835&0.976&0.935&\textbf{0.771}\\
      DoubleU-Net \cite{jha2020doubleu}& 0.845 & 0.841&0.967&0.933&0.760\\
      TransUnet \cite{chen2021transunet}& 0.841 & 0.807&0.979&0.932&0.755\\
      FAT-Net \cite{wu2022fat}& 0.850 & 0.840&0.973&0.933&0.765\\
      Swin-Unet \cite{cao2021swin}& 0.862 & 0.842&0.950&0.936&0.764\\
      M2SNet \cite{zhao2023m}& 0.853 & 0.828&0.964&0.931&0.751\\
      UTNetV2 \cite{gao2022multi}& 0.831 & 0.791&0.972&0.927&0.742\\
      MISSFormer \cite{huang2021missformer}& 0.863 & 0.848&0.980&0.939&0.764\\
      \midrule
      Ours(BEFUnet) & \textbf{0.868} & \textbf{0.853}&\textbf{0.985}&\textbf{0.946}&0.768\\
      \hline
    \end{tabular*}
  \end{center}
\end{table}

\begin{figure*}[htb]
    \centering
    \includegraphics[width=0.8\textwidth]{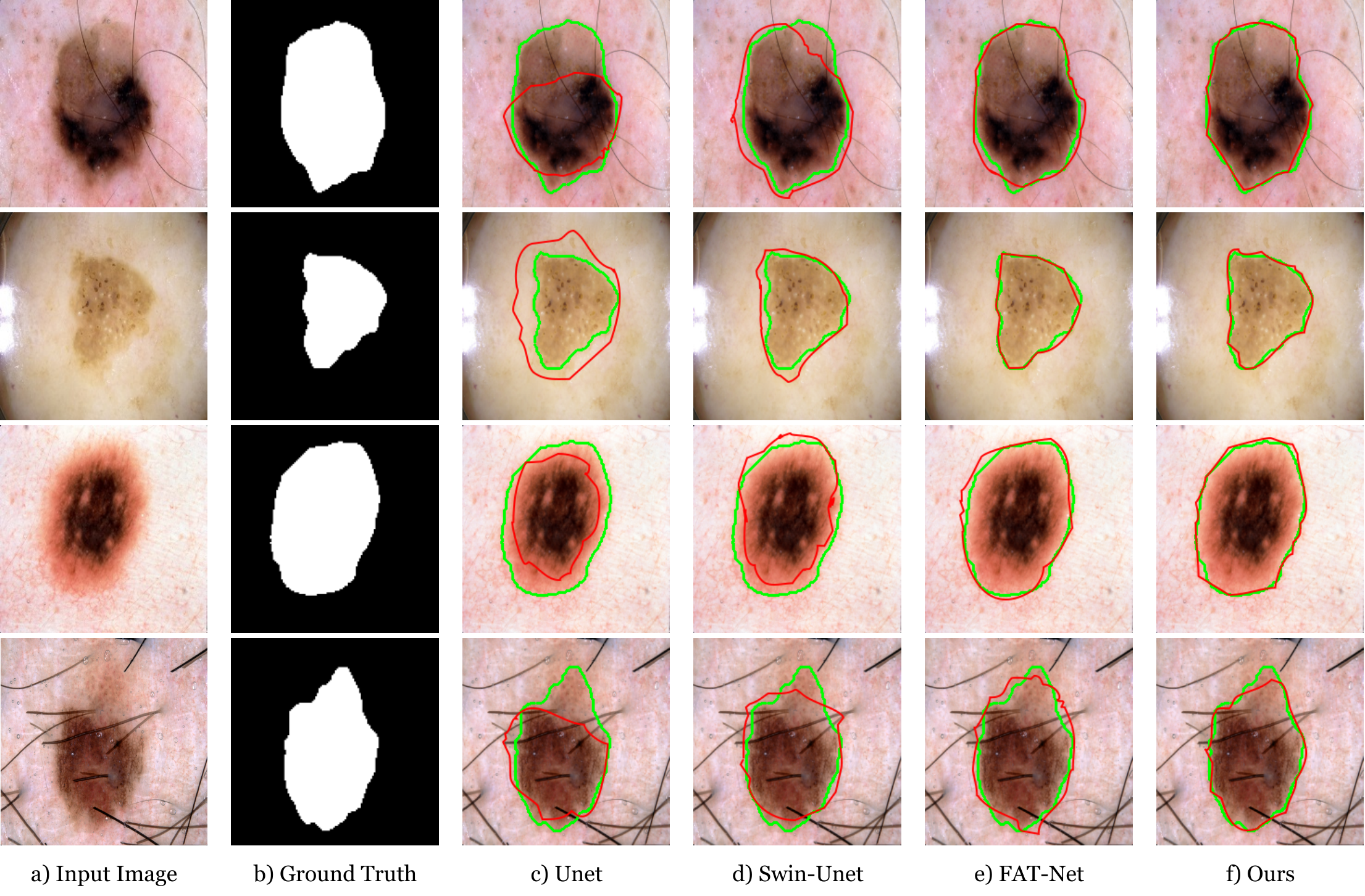}

    \vspace{-0.75em}
    \caption{Visual comparison with different SOTA methods on the ISIC 2017 dataset. The \textcolor{green}{green} contours are ground truth, and the \textcolor{red}{red} contours are the segmentation results of indicated methods.}
    \label{fig:skin}
\end{figure*}

\subsection{Results of Skin Lesion Segmentation}

Table \ref{Tab:ISIC} displays the comparison results for the ISIC 2017 benchmarks. Our method stands out with the highest scores for most metrics, achieving 86.8\% for Dice, 85.3\% for SE, 98.5\% for SP, 94.6\% for ACC, , and 76.8\% for IoU.

Additionally, we visually compared the segmentation results of several representative competitors in our experiments, namely U-Net, Swin-Net, and FAT-Net (Figure \ref{fig:skin}). The observations reveal that our method consistently outperforms the competition, particularly in challenging examples. U-Net struggles to accurately predict skin lesions when there is low contrast between foreground and background pixels due to the lack of sufficient global contextual information. FAT-Net, by incorporating a pyramidal module for multi-scale feature extraction, mitigates some of the weaknesses in global context information extraction, but it still produces mis-segmentation in complex boundary cases.
In contrast, our BEFUnet successfully integrates multi-scale encoders to capture both edge features and body context information. This enables us to achieve superior segmentation results compared to other competitors, especially in challenging scenarios involving ambiguous brightness contrast and boundary situations.

\begin{table*}[t!]
\centering
\caption{The ablation experiment results of BEFUnet on the Synapse dataset.}
\footnotesize
\resizebox{\textwidth}{!}{
\begin{tabular}{c|cc|cccccccc}
\hline
\textbf{Method} &  DSC$\uparrow$ &HD$\downarrow$ & Aorta& Gallbladder& Kidney(L)& Kidney(R)& Liver& Pancreas& Spleen& Stomach   \\
\hline
Baseline &             70.19  & 30.88  & 80.70  & 67.08  & 72.85  & 65.03  & 89.92  & 55.53  & 82.22  & 68.18  \\
Baseline + DLF &       71.72  & 27.97  & 81.59  & 68.73  & 75.05  & 67.96  & 91.15  & 56.03  & 84.13  & 69.09  \\
Baseline + EE &        74.54  & 25.84  & 84.98  & 69.27  & 75.43  & 69.16  & 92.53  & 57.20  & 87.18  & 73.58  \\
Baseline + EE + LCAF & 78.91  & 20.86  & 86.89  & 71.01  & 81.57  & 76.18  & 92.86  & 58.14  & 87.62  & 75.19  \\
Baseline + EE + LCAF + DLF (BEFUnet) & \textbf{80.47} &  \textbf{16.26} & \textbf{87.03} & \textbf{73.89} & \textbf{85.23} & \textbf{80.47} & \textbf{95.49} & \textbf{60.31} & \textbf{91.56} & \textbf{81.54} \\
\hline
\end{tabular}
}
\label{synapse_2}
\end{table*}

\section{Ablation Study}
\noindent\textbf{Efficiency of Different Components}
Ablation experiments were performed to showcase the effectiveness of various components within the LCAUnet model. The baseline configuration consists of a single-branch U-shaped architecture comprising a body encoder and a decoder, utilizing a straightforward concatenation of adjacent-scale features. Subsequently, the baseline network was enhanced by individually incorporating the LCAF module, Edge-Encoder (EE), and DLF module, resulting in three distinct methods denoted as baseline+DLF, baseline+EE, and baseline+EE+LCAF. Comparative evaluations were carried out on these modified networks with different modules using the Synapse dataset.

In Table \ref{synapse_2}, we present a comparison between the baseline method and two augmented methods: baseline+DLF and baseline+EE. The results demonstrate notable improvements for both augmented methods on the Synapse dataset. Specifically, the baseline+DLF method shows a remarkable enhancement of approximately 2.91\% in the HD metric and approximately 1.53\% in the DSC metric. Similarly, the baseline+EE method achieves significant improvements, with approximately 5.04\% lower HD and approximately 4.35\% higher DSC.
These results underscore the importance of the DLF module, as it effectively facilitates the fusion of features across adjacent scales. Additionally, the Edge-Encoder proves to be crucial in extracting edge information, leading to enhanced segmentation performance.
Furthermore, the combination of baseline+EE+LCAF method offers even better results, with a further improve of 4.98\% in the HD metric and 4.37\% in the DSC metric compared to the baseline+EE method. This further highlights the valuable role played by the LCAF module in promoting the fusion of edge and body information, contributing to improved performance.
Ultimately, by integrating all three modules into our comprehensive approach (baseline+EE+LCAF+DLF), we achieve the best performance across all metrics. Specifically, this approach surpasses the baseline method by approximately 10.28\% in DSC, and 14.62\% in HD metrics on the Synapse dataset.


\section{Discussion}
Our extensive experiments conducted on various medical image segmentation datasets showcase the effectiveness of our novel BEFUnet model when compared to traditional CNN and transformer-based approaches. Our approach encompasses two key advancements. To enhance the design rationality of the network, we have incorporated Pixel Different Convolution and Transformer into the initial layers. Furthermore, the LCAF module facilitates feature reuse by merging the Edge features from PDC with the body features provided by the transformer module. The quantitative evaluation of the BEFUnet network on three challenging datasets demonstrates its exceptional segmentation performance, outperforming SOTA methods in most cases. This result is also supported by visual analysis, as depicted in Fig. \ref{fig:synapseviz}, which shows noise-free segmentation of organs such as the Kidney and Liver, consistent with the quantitative benchmarks. However, our model may encounter failure cases in certain instances (e.g., Aorta), aligning with the numerical results. Additionally, our model still faces notable challenges when dealing with low-contrast skin images. Overall, BEFUnet has demonstrated the ability to effectively learn critical anatomical relationships from medical images.
%

\section{Conclusions}
this paper makes three significant contributions. Firstly, we introduce the concept of pixel difference convolution, which combines the strengths of traditional edge encoders and deep CNNs. This integration results in robust and precise edge segmentation. Secondly, we propose a novel medical image segmentation structure called BEFUnet, which extracts both body and edge features and effectively integrates them to enhance segmentation performance. To achieve this, we construct a local cross-modal fusion module, LCAF, that combines body and edge modalities for feature fusion. Furthermore, we employ a DLF module to achieve finer feature fusion from the aforementioned representations. Lastly, we conduct comprehensive experiments on three publicly available datasets to evaluate the proposed model. Compared to existing methods, the BEFUnet model demonstrates superior performance. It accurately segments boundary-blurred, irregular, and interference-present organs regions, while exhibiting strong generalization capabilities. The results validate that the proposed BEFUnet model outperforms most SOTA methods.

{\small
\bibliographystyle{ieee_fullname}
\bibliography{egbib}

\begin{thebibliography}{10}\itemsep=-1pt

\bibitem{alom2018recurrent}
Md~Zahangir Alom, Mahmudul Hasan, Chris Yakopcic, Tarek~M Taha, and Vijayan~K
  Asari.
\newblock Recurrent residual convolutional neural network based on u-net
  (r2u-net) for medical image segmentation.
\newblock {\em arXiv preprint arXiv:1802.06955}, 2018.

\bibitem{badrinarayanan2017segnet}
Vijay Badrinarayanan, Alex Kendall, and Roberto Cipolla.
\newblock Segnet: A deep convolutional encoder-decoder architecture for image
  segmentation.
\newblock {\em IEEE transactions on pattern analysis and machine intelligence},
  39(12):2481--2495, 2017.

\bibitem{bai2022transfusion}
Xuyang Bai, Zeyu Hu, Xinge Zhu, Qingqiu Huang, Yilun Chen, Hongbo Fu, and
  Chiew-Lan Tai.
\newblock Transfusion: Robust lidar-camera fusion for 3d object detection with
  transformers.
\newblock In {\em Proceedings of the IEEE/CVF conference on computer vision and
  pattern recognition}, pages 1090--1099, 2022.

\bibitem{cai2020dense}
Sijing Cai, Yunxian Tian, Harvey Lui, Haishan Zeng, Yi Wu, and Guannan Chen.
\newblock Dense-unet: a novel multiphoton in vivo cellular image segmentation
  model based on a convolutional neural network.
\newblock {\em Quantitative imaging in medicine and surgery}, 10(6):1275, 2020.

\bibitem{cao2021swin}
Hu Cao, Yueyue Wang, Joy Chen, Dongsheng Jiang, Xiaopeng Zhang, Qi Tian, and
  Manning Wang.
\newblock Swin-unet: Unet-like pure transformer for medical image segmentation.
\newblock {\em arXiv preprint arXiv:2105.05537}, 2021.

\bibitem{synapse2015ct}
MICCAI 2015 Multi-Atlas Abdomen~Labeling Challenge.
\newblock Synapse multi-organ segmentation dataset.
\newblock https://www.synapse.org/\#!Synapse:syn3193805/wiki/217789, 2015.
\newblock Accessed: 2022-04-20.

\bibitem{chen2021crossvit}
Chun-Fu~Richard Chen, Quanfu Fan, and Rameswar Panda.
\newblock Crossvit: Cross-attention multi-scale vision transformer for image
  classification.
\newblock In {\em Proceedings of the IEEE/CVF International Conference on
  Computer Vision}, pages 357--366, 2021.

\bibitem{chen2021transunet}
Jieneng Chen, Yongyi Lu, Qihang Yu, Xiangde Luo, Ehsan Adeli, Yan Wang, Le Lu,
  Alan~L Yuille, and Yuyin Zhou.
\newblock Transunet: Transformers make strong encoders for medical image
  segmentation.
\newblock {\em arXiv preprint arXiv:2102.04306}, 2021.

\bibitem{chen2017deeplab}
Liang-Chieh Chen, George Papandreou, Iasonas Kokkinos, Kevin Murphy, and Alan~L
  Yuille.
\newblock Deeplab: Semantic image segmentation with deep convolutional nets,
  atrous convolution, and fully connected crfs.
\newblock {\em IEEE transactions on pattern analysis and machine intelligence},
  40(4):834--848, 2017.

\bibitem{chen2018encoder}
Liang-Chieh Chen, Yukun Zhu, George Papandreou, Florian Schroff, and Hartwig
  Adam.
\newblock Encoder-decoder with atrous separable convolution for semantic image
  segmentation.
\newblock In {\em Proceedings of the European conference on computer vision
  (ECCV)}, pages 801--818, 2018.

\bibitem{cheng2020learning}
Feng Cheng, Cheng Chen, Yukang Wang, Heshui Shi, Yukun Cao, Dandan Tu,
  Changzheng Zhang, and Yongchao Xu.
\newblock Learning directional feature maps for cardiac mri segmentation.
\newblock In {\em Medical Image Computing and Computer Assisted
  Intervention--MICCAI 2020: 23rd International Conference, Lima, Peru, October
  4--8, 2020, Proceedings, Part IV 23}, pages 108--117. Springer, 2020.

\bibitem{codella2019skin}
Noel Codella, Veronica Rotemberg, Philipp Tschandl, M~Emre Celebi, Stephen
  Dusza, David Gutman, Brian Helba, Aadi Kalloo, Konstantinos Liopyris, Michael
  Marchetti, et~al.
\newblock Skin lesion analysis toward melanoma detection 2018: A challenge
  hosted by the international skin imaging collaboration (isic).
\newblock {\em arXiv preprint arXiv:1902.03368}, 2019.

\bibitem{codella2018skin}
Noel~CF Codella, David Gutman, M~Emre Celebi, Brian Helba, Michael~A Marchetti,
  Stephen~W Dusza, Aadi Kalloo, Konstantinos Liopyris, Nabin Mishra, Harald
  Kittler, et~al.
\newblock Skin lesion analysis toward melanoma detection: A challenge at the
  2017 international symposium on biomedical imaging (isbi), hosted by the
  international skin imaging collaboration (isic).
\newblock In {\em 2018 IEEE 15th international symposium on biomedical imaging
  (ISBI 2018)}, pages 168--172. IEEE, 2018.

\bibitem{dosovitskiy2020image}
Alexey Dosovitskiy, Lucas Beyer, Alexander Kolesnikov, Dirk Weissenborn,
  Xiaohua Zhai, Thomas Unterthiner, Mostafa Dehghani, Matthias Minderer, Georg
  Heigold, Sylvain Gelly, et~al.
\newblock An image is worth 16x16 words: Transformers for image recognition at
  scale.
\newblock {\em arXiv preprint arXiv:2010.11929}, 2020.

\bibitem{fu2020domain}
Shuhao Fu, Yongyi Lu, Yan Wang, Yuyin Zhou, Wei Shen, Elliot Fishman, and Alan
  Yuille.
\newblock Domain adaptive relational reasoning for 3d multi-organ segmentation.
\newblock In {\em International Conference on Medical Image Computing and
  Computer-Assisted Intervention}, pages 656--666. Springer, 2020.

\bibitem{gao2022multi}
Yunhe Gao, Mu Zhou, Di Liu, and D Metaxas.
\newblock A multi-scale transformer for medical image segmentation:
  Architectures, model efficiency, and benchmarks.
\newblock {\em arXiv preprint arXiv:2203.00131}, 2022.

\bibitem{gao2022data}
Yunhe Gao, Mu Zhou, Di Liu, Zhennan Yan, Shaoting Zhang, and Dimitris~N
  Metaxas.
\newblock A data-scalable transformer for medical image segmentation:
  architecture, model efficiency, and benchmark.
\newblock {\em arXiv preprint arXiv:2203.00131}, 2022.

\bibitem{gu2022multi}
Jiaqi Gu, Hyoukjun Kwon, Dilin Wang, Wei Ye, Meng Li, Yu-Hsin Chen, Liangzhen
  Lai, Vikas Chandra, and David~Z Pan.
\newblock Multi-scale high-resolution vision transformer for semantic
  segmentation.
\newblock In {\em Proceedings of the IEEE/CVF Conference on Computer Vision and
  Pattern Recognition}, pages 12094--12103, 2022.

\bibitem{segpc2021}
Anubha Gupta, Ritu Gupta, Shiv Gehlot, and Shubham Goswami.
\newblock Segpc-2021: Segmentation of multiple myeloma plasma cells in
  microscopic images, 2021.

\bibitem{gupta2018pcseg}
Anubha Gupta, Pramit Mallick, Ojaswa Sharma, Ritu Gupta, and Rahul Duggal.
\newblock Pcseg: Color model driven probabilistic multiphase level set based
  tool for plasma cell segmentation in multiple myeloma.
\newblock {\em PloS one}, 13(12):e0207908, 2018.

\bibitem{huang2020unet}
Huimin Huang, Lanfen Lin, Ruofeng Tong, Hongjie Hu, Qiaowei Zhang, Yutaro
  Iwamoto, Xianhua Han, Yen-Wei Chen, and Jian Wu.
\newblock Unet 3+: A full-scale connected unet for medical image segmentation.
\newblock In {\em ICASSP 2020-2020 IEEE International Conference on Acoustics,
  Speech and Signal Processing (ICASSP)}, pages 1055--1059. IEEE, 2020.

\bibitem{huang2021missformer}
Xiaohong Huang, Zhifang Deng, Dandan Li, and Xueguang Yuan.
\newblock Missformer: An effective medical image segmentation transformer.
\newblock {\em arXiv preprint arXiv:2109.07162}, 2021.

\bibitem{jha2020doubleu}
Debesh Jha, Michael~A Riegler, Dag Johansen, P{\aa}l Halvorsen, and
  H{\aa}vard~D Johansen.
\newblock Doubleu-net: A deep convolutional neural network for medical image
  segmentation.
\newblock In {\em 2020 IEEE 33rd International symposium on computer-based
  medical systems (CBMS)}, pages 558--564. IEEE, 2020.

\bibitem{jha2019resunet++}
Debesh Jha, Pia~H Smedsrud, Michael~A Riegler, Dag Johansen, Thomas De~Lange,
  P{\aa}l Halvorsen, and H{\aa}vard~D Johansen.
\newblock Resunet++: An advanced architecture for medical image segmentation.
\newblock In {\em 2019 IEEE international symposium on multimedia (ISM)}, pages
  225--2255. IEEE, 2019.

\bibitem{kaul2019focusnet}
Chaitanya Kaul, Suresh Manandhar, and Nick Pears.
\newblock Focusnet: An attention-based fully convolutional network for medical
  image segmentation.
\newblock In {\em 2019 IEEE 16th international symposium on biomedical imaging
  (ISBI 2019)}, pages 455--458. IEEE, 2019.

\bibitem{kuang2021bea}
Hulin Kuang, Yixiong Liang, Ning Liu, Jin Liu, and Jianxin Wang.
\newblock Bea-segnet: Body and edge aware network for medical image
  segmentation.
\newblock In {\em 2021 IEEE International Conference on Bioinformatics and
  Biomedicine (BIBM)}, pages 939--944. IEEE, 2021.

\bibitem{lei2020skin}
Baiying Lei, Zaimin Xia, Feng Jiang, Xudong Jiang, Zongyuan Ge, Yanwu Xu, Jing
  Qin, Siping Chen, Tianfu Wang, and Shuqiang Wang.
\newblock Skin lesion segmentation via generative adversarial networks with
  dual discriminators.
\newblock {\em Medical Image Analysis}, 64:101716, 2020.

\bibitem{li2022deepfusion}
Yingwei Li, Adams~Wei Yu, Tianjian Meng, Ben Caine, Jiquan Ngiam, Daiyi Peng,
  Junyang Shen, Yifeng Lu, Denny Zhou, Quoc~V Le, et~al.
\newblock Deepfusion: Lidar-camera deep fusion for multi-modal 3d object
  detection.
\newblock In {\em Proceedings of the IEEE/CVF Conference on Computer Vision and
  Pattern Recognition}, pages 17182--17191, 2022.

\bibitem{lin2022ds}
Ailiang Lin, Bingzhi Chen, Jiayu Xu, Zheng Zhang, Guangming Lu, and David
  Zhang.
\newblock Ds-transunet: Dual swin transformer u-net for medical image
  segmentation.
\newblock {\em IEEE Transactions on Instrumentation and Measurement}, 2022.

\bibitem{liu2017richer}
Yun Liu, Ming-Ming Cheng, Xiaowei Hu, Kai Wang, and Xiang Bai.
\newblock Richer convolutional features for edge detection.
\newblock In {\em Proceedings of the IEEE conference on computer vision and
  pattern recognition}, pages 3000--3009, 2017.

\bibitem{liu2021swin}
Ze Liu, Yutong Lin, Yue Cao, Han Hu, Yixuan Wei, Zheng Zhang, Stephen Lin, and
  Baining Guo.
\newblock Swin transformer: Hierarchical vision transformer using shifted
  windows.
\newblock In {\em Proceedings of the IEEE/CVF International Conference on
  Computer Vision}, pages 10012--10022, 2021.

\bibitem{long2015fully}
Jonathan Long, Evan Shelhamer, and Trevor Darrell.
\newblock Fully convolutional networks for semantic segmentation.
\newblock In {\em Proceedings of the IEEE conference on computer vision and
  pattern recognition}, pages 3431--3440, 2015.

\bibitem{mendoncca2013ph}
Teresa Mendon{\c{c}}a, Pedro~M Ferreira, Jorge~S Marques, Andr{\'e}~RS Marcal,
  and Jorge Rozeira.
\newblock Ph 2-a dermoscopic image database for research and benchmarking.
\newblock In {\em 2013 35th annual international conference of the IEEE
  engineering in medicine and biology society (EMBC)}, pages 5437--5440. IEEE,
  2013.

\bibitem{milletari2016v}
Fausto Milletari, Nassir Navab, and Seyed-Ahmad Ahmadi.
\newblock V-net: Fully convolutional neural networks for volumetric medical
  image segmentation.
\newblock In {\em 2016 fourth international conference on 3D vision (3DV)},
  pages 565--571. Ieee, 2016.

\bibitem{oktay2018attention}
Ozan Oktay, Jo Schlemper, Loic~Le Folgoc, Matthew Lee, Mattias Heinrich,
  Kazunari Misawa, Kensaku Mori, Steven McDonagh, Nils~Y Hammerla, Bernhard
  Kainz, et~al.
\newblock Attention u-net: Learning where to look for the pancreas.
\newblock {\em arXiv preprint arXiv:1804.03999}, 2018.

\bibitem{ronneberger2015u}
Olaf Ronneberger, Philipp Fischer, and Thomas Brox.
\newblock U-net: Convolutional networks for biomedical image segmentation.
\newblock In {\em International Conference on Medical image computing and
  computer-assisted intervention}, pages 234--241. Springer, 2015.

\bibitem{ruan2023ege}
Jiacheng Ruan, Mingye Xie, Jingsheng Gao, Ting Liu, and Yuzhuo Fu.
\newblock Ege-unet: an efficient group enhanced unet for skin lesion
  segmentation.
\newblock In {\em International Conference on Medical Image Computing and
  Computer-Assisted Intervention}, pages 481--490. Springer, 2023.

\bibitem{schlemper2019attention}
Jo Schlemper, Ozan Oktay, Michiel Schaap, Mattias Heinrich, Bernhard Kainz, Ben
  Glocker, and Daniel Rueckert.
\newblock Attention gated networks: Learning to leverage salient regions in
  medical images.
\newblock {\em Medical image analysis}, 53:197--207, 2019.

\bibitem{su2021pixel}
Zhuo Su, Wenzhe Liu, Zitong Yu, Dewen Hu, Qing Liao, Qi Tian, Matti
  Pietik{\"a}inen, and Li Liu.
\newblock Pixel difference networks for efficient edge detection.
\newblock In {\em Proceedings of the IEEE/CVF international conference on
  computer vision}, pages 5117--5127, 2021.

\bibitem{touvron2021training}
Hugo Touvron, Matthieu Cord, Matthijs Douze, Francisco Massa, Alexandre
  Sablayrolles, and Herv{\'e} J{\'e}gou.
\newblock Training data-efficient image transformers \& distillation through
  attention.
\newblock In {\em International Conference on Machine Learning}, pages
  10347--10357. PMLR, 2021.

\bibitem{tsai2003shape}
Andy Tsai, Anthony Yezzi, William Wells, Clare Tempany, Dewey Tucker, Ayres
  Fan, W~Eric Grimson, and Alan Willsky.
\newblock A shape-based approach to the segmentation of medical imagery using
  level sets.
\newblock {\em IEEE transactions on medical imaging}, 22(2):137--154, 2003.

\bibitem{valanarasu2021medical}
Jeya Maria~Jose Valanarasu, Poojan Oza, Ilker Hacihaliloglu, and Vishal~M
  Patel.
\newblock Medical transformer: Gated axial-attention for medical image
  segmentation.
\newblock In {\em Medical Image Computing and Computer Assisted
  Intervention--MICCAI 2021: 24th International Conference, Strasbourg, France,
  September 27--October 1, 2021, Proceedings, Part I 24}, pages 36--46.
  Springer, 2021.

\bibitem{valanarasu2020kiu}
Jeya Maria~Jose Valanarasu, Vishwanath~A Sindagi, Ilker Hacihaliloglu, and
  Vishal~M Patel.
\newblock Kiu-net: Towards accurate segmentation of biomedical images using
  over-complete representations.
\newblock In {\em International conference on medical image computing and
  computer-assisted intervention}, pages 363--373. Springer, 2020.

\bibitem{vaswani2017attention}
Ashish Vaswani, Noam Shazeer, Niki Parmar, Jakob Uszkoreit, Llion Jones,
  Aidan~N Gomez, {\L}ukasz Kaiser, and Illia Polosukhin.
\newblock Attention is all you need.
\newblock {\em Advances in neural information processing systems}, 30, 2017.

\bibitem{wu2022fat}
Huisi Wu, Shihuai Chen, Guilian Chen, Wei Wang, Baiying Lei, and Zhenkun Wen.
\newblock Fat-net: Feature adaptive transformers for automated skin lesion
  segmentation.
\newblock {\em Medical Image Analysis}, 76:102327, 2022.

\bibitem{xiao2023transformers}
Hanguang Xiao, Li Li, Qiyuan Liu, Xiuhong Zhu, and Qihang Zhang.
\newblock Transformers in medical image segmentation: A review.
\newblock {\em Biomedical Signal Processing and Control}, 84:104791, 2023.

\bibitem{xiao2018weighted}
Xiao Xiao, Shen Lian, Zhiming Luo, and Shaozi Li.
\newblock Weighted res-unet for high-quality retina vessel segmentation.
\newblock In {\em 2018 9th international conference on information technology
  in medicine and education (ITME)}, pages 327--331. IEEE, 2018.

\bibitem{xie2015holistically}
Saining Xie and Zhuowen Tu.
\newblock Holistically-nested edge detection.
\newblock In {\em Proceedings of the IEEE international conference on computer
  vision}, pages 1395--1403, 2015.

\bibitem{xu2021levit}
Guoping Xu, Xingrong Wu, Xuan Zhang, and Xinwei He.
\newblock Levit-unet: Make faster encoders with transformer for medical image
  segmentation.
\newblock {\em arXiv preprint arXiv:2107.08623}, 2021.

\bibitem{yang2023cswin}
Haonan Yang and Dapeng Yang.
\newblock Cswin-pnet: A cnn-swin transformer combined pyramid network for
  breast lesion segmentation in ultrasound images.
\newblock {\em Expert Systems with Applications}, 213:119024, 2023.

\bibitem{yu2015multi}
Fisher Yu and Vladlen Koltun.
\newblock Multi-scale context aggregation by dilated convolutions.
\newblock {\em arXiv preprint arXiv:1511.07122}, 2015.

\bibitem{zhao2017pyramid}
Hengshuang Zhao, Jianping Shi, Xiaojuan Qi, Xiaogang Wang, and Jiaya Jia.
\newblock Pyramid scene parsing network.
\newblock In {\em Proceedings of the IEEE conference on computer vision and
  pattern recognition}, pages 2881--2890, 2017.

\bibitem{zhao2023m}
Xiaoqi Zhao, Hongpeng Jia, Youwei Pang, Long Lv, Feng Tian, Lihe Zhang, Weibing
  Sun, and Huchuan Lu.
\newblock Msnet: Multi-scale in multi-scale subtraction network for medical
  image segmentation.
\newblock {\em arXiv preprint arXiv:2303.10894}, 2023.

\bibitem{zhou2018unet++}
Zongwei Zhou, Md~Mahfuzur Rahman~Siddiquee, Nima Tajbakhsh, and Jianming Liang.
\newblock Unet++: A nested u-net architecture for medical image segmentation.
\newblock In {\em Deep learning in medical image analysis and multimodal
  learning for clinical decision support}, pages 3--11. Springer, 2018.

\bibitem{zhu2023brain}
Zhiqin Zhu, Xianyu He, Guanqiu Qi, Yuanyuan Li, Baisen Cong, and Yu Liu.
\newblock Brain tumor segmentation based on the fusion of deep semantics and
  edge information in multimodal mri.
\newblock {\em Information Fusion}, 91:376--387, 2023.

\end{thebibliography}
}

\end{document}